\documentclass[final,3p,times,twocolumn]{elsarticle}

\usepackage{amssymb}

\usepackage{latexsym}
\usepackage{amsmath}
\usepackage{multirow}

\begin{document}

\begin{frontmatter}


\author[1]{Hariom A. Pandya}
\author[1]{Dr. Brijesh S. Bhatt}
 \affiliation[1]{organization={Computer Engineering Department},
             addressline={Dharmsinh Desai University},
             city={Nadiad},
             state={Gujarat},
             country={India}}

\title{Question Answering Survey: Directions, Challenges, Datasets, Evaluation Matrices}


\begin{abstract}
The usage and amount of information available on the internet increase over the past decade. This digitization leads to the need for automated answering system to extract fruitful information from redundant and transitional knowledge sources. Such systems are designed to cater the most prominent answer from this giant knowledge source to the user’s query using natural language understanding (NLU) and thus eminently depends on the Question-answering(QA) field.

Question answering involves but not limited to the steps like mapping of user’s question to pertinent query, retrieval of relevant information, finding the best suitable answer from the retrieved information etc. The current improvement of deep learning models evince compelling performance improvement in all these tasks.

In this review work, the research directions of QA field are analyzed based on the type of question, answer type, source of evidence-answer, and modeling approach. This detailing followed by open challenges of the field like automatic question generation, similarity detection and, low resource availability for a language. In the end, a survey of available datasets and evaluation measures is presented.
\end{abstract}




\end{frontmatter}


\section{Introduction}

Information retrieval(IR) has been an active area of research for past many decades as manual retrieval of fruitful information from enormous data is a tedious and time-consuming task. Handling this issue and retrieval of updated and important information brings the attention of many researchers \cite{wang2015long,zheng2002answerbus,cui2005question,kwok2001scaling}. The notable performance improvement in the area can be observed by digital personal assistants (such as Amazon Alexa, Apple Siri, Google Home, etc.), robots-communication, clinical uses of conversation agents for mental health, Chatbot, etc.

Generally, these applications are tuned to have users input in the form of text, video, or speech \cite{zhang2018variational,wang2017gated,braun2017evaluating,yu2019deep,moses2019real,dave2020satellite,diefenbach2018core,radford2018improving,lei-etal-2018-tvqa}. Based on the question or input, the system will perform the required action. The pivotal requirement here is to understand and process the output produced by the information retrieval(IR) module \cite{yang2019end,horvitz2009system}, and is the role of the question answering system.

Question answering is a field of information retrieval\cite{kwok2001scaling} and natural language processing (NLP), concerned with building systems that automatically answer questions posed by humans in a natural language.\footnote[1]{The definition is borrowed from the Wikipedia. https://en.wikipedia.org/wiki/Question\_answering}.

\begin{figure}[h]
\centering
 \includegraphics[width=0.4\textwidth,keepaspectratio]{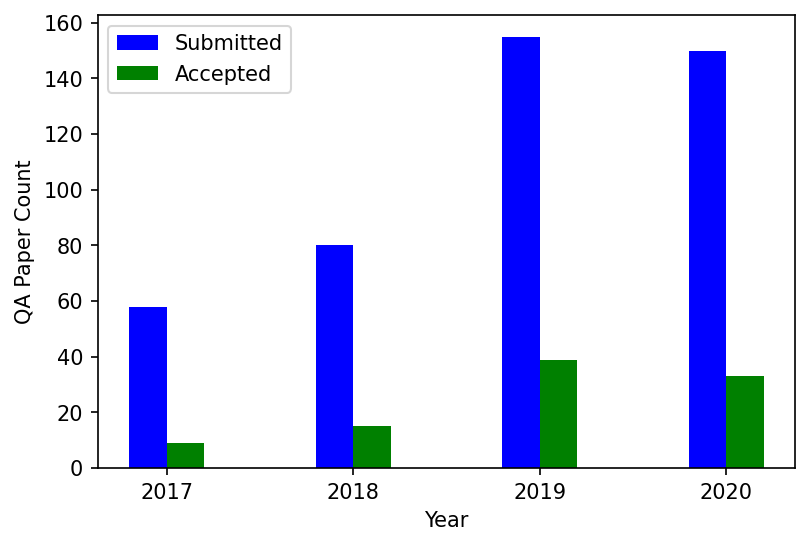}
 \caption{Count of submitted and accepted QA paper in Association for Computational Linguistics(ACL)}
 \label{fig:Year_VS_QA_Papers} 
\end{figure}

As many syntax, semantic, and discourse level challenges involved in this field, it is one of the prominent research directions of recent decades. This can be witnessed by figure \ref{fig:Year_VS_QA_Papers} which indicates statistics of total submitted and accepted papers of QA field in Association for Computational Linguistics (ACL). It clearly indicates the surge in the popularity of the field. 

Pragmatically helpful resources for answering the query include Wikipedia, quora, Reddit, tweeter, stackoverflow, etc.\cite{rajpurkar-etal-2016-squad,rajpurkar-etal-2018-know,yang-etal-2015-wikiqa,1162-natural,10.5555/2969239.2969428,fan-etal-2018-hierarchical,joshi-etal-2017-triviaqa,dhingra2017quasar,yu2018modelling}. Many researchers are working in the applications which entertain users query by identifying desired information from such colossal datasets and cater it in accordance to the query.

Finding the best possible answer from one of these datasets is a challenging task \cite{10.1145/2970398.2970438,10.1145/3357384.3358026,ravichandran2002learning,voorhees1999trec} due to features selection, its mapping with model and finding relevancy ranking of all candidates answers etc. common challenges. Apart from this, there are some application-specific key challenges involved in question answering tasks, some of them are highlighted in this paper.

Some of the early notable work of the IR tasks are the BASEBALL and the LUNAR system. BASEBALL is a program developed by Green et al. \cite{10.1145/1460690.1460714}, which is considered as the foundation of QA field, by many researchers. The system was aimed to answer questions about baseball game played in the American League. The idea there is to fetch the answer, stored in a punch card, with the help of a dictionary. The next breakthrough of the QA field is in 1971, LUNAR System(Lunar Sciences Natural Language Information System) \cite{10.1145/1499586.1499695}, a system designed for chemical analysis of Lunar rocks. It contains database files provided by NASA with 13,000 entry tables of chemical, isotope, and age analyses of the Apollo 11 samples.

Starting from such Information Retrieval (IR) tasks to current encoder-decoder based question-answering, the field has witnessed tremendous progress especially after the involvement of machine learning and deep learning techniques in the field \cite{Reddy2019,chen-etal-2017-reading,wang2015long,zheng2002answerbus,cui2005question,kwok2001scaling,huang2019knowledge,Lv_Guo_Xu_Tang_Duan_Gong_Shou_Jiang_Cao_Hu_2020}. The availability of high computing machines is another key factor for this exponential advancement.

\begin{figure}[h]
\centering
 \includegraphics[width=0.4\textwidth,keepaspectratio]{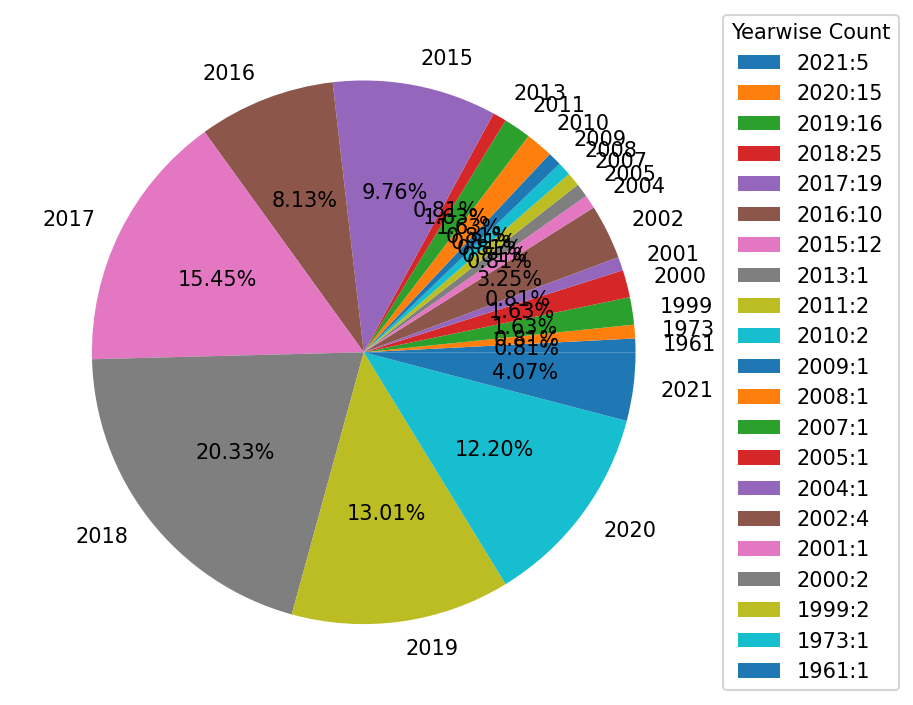}
 \caption{Year wise Surveyed Papers}
 \label{fig:paperPerYear} 
\end{figure}

This survey covers advancement of the QA field with predominant focus on the deep learning based QA approaches. The year wise distribution of the surveyed papers is mentioned in Figure \ref{fig:paperPerYear}.
 
The remainder of this paper is organized as follows: In the next section, we explore the QA categorization based on the parameters like types of question, answer type, source of evidence or answer, and modeling approaches of the field. Next, we explore the machine reading comprehension (MRC) datasets and Knowledge graph(KB) based datasets followed by evaluation matrices of QA field. In the end observations with statistics are catered.

\section{Categorization of QA system}

\begin{figure*}
\centering
 \includegraphics[width=\textwidth,keepaspectratio]{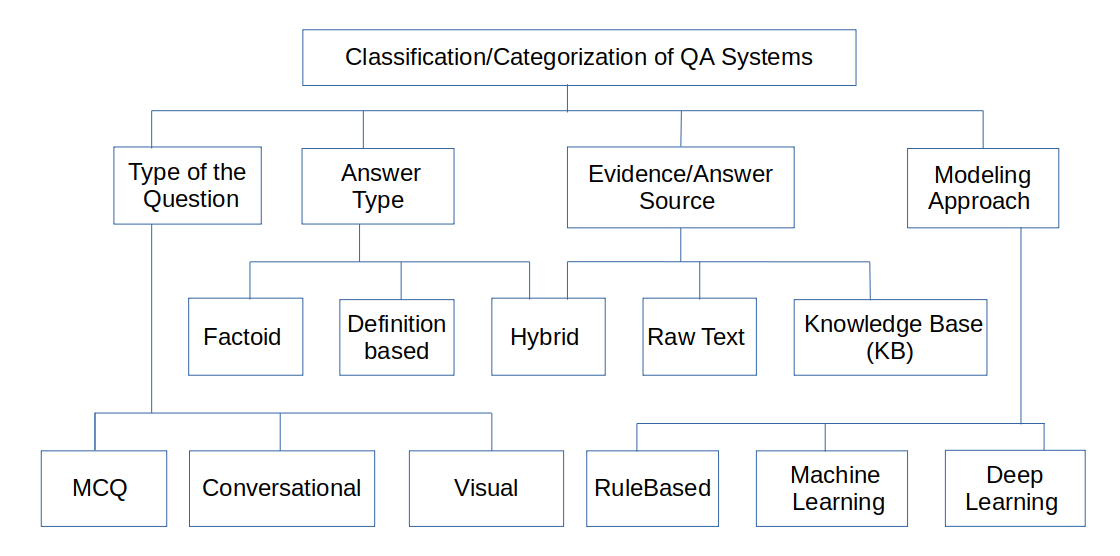}
 \caption{Question Answering : Categories}
 \label{fig:category} 
\end{figure*}

Research in the field of Question Answering is diversified due to the reasons like type of question, expected answer type, source of answer evidence, answer retrieval modeling approach etc. In this study we have covered these categories as mentioned in the Figure \ref{fig:category}. In further subsections survey of each is catered. The table \ref{table1} indicates summary of individual category used in the selected papers between years 2017-20.

\begin{table}[!htbp]
    \centering
    
    \resizebox{\columnwidth}{!}{%
    \begin{tabular}{p{0.48\textwidth}|c|c|c|c|c|c}
     \hline
     \textbf{Paper} & \textbf{Year} & 
    \textbf{QT} & \textbf{AT} & 
    \textbf{ES} & \textbf{AS} & 
    \textbf{MA}\\
    \hline
     Reading Wikipedia to Answer Open-Domain Questions \cite{chen-etal-2017-reading} & 2017 & RC & Def & RT & RT & DL \\
     \hline
     Leveraging video descriptions to learn video question answering \cite{10.5555/3298023.3298196} & 2017 & Visual & Fact & Hybrid & Hybrid & DL \\
     \hline
    
    End to end long short term memory networks for non-factoid question answering  \cite{10.1145/2970398.2970438} & 2017 & RC & Def & RT & RT & DL \\
     \hline
     
    Improving Deep Learning for Multiple Choice Question Answering with Candidate Contexts \cite{nicula2018improving} & 2018 & MCQ & Def & RT & RT & DL \\
     \hline

    A Better Way to Attend: Attention with Trees for Video Question Answering  \cite{8419716} & 2018 & Visual & Fact & Hybrid & Hybrid & DL \\
    
     \hline
     Knowledge base question answering via encoding of complex query graphs \cite{luo2018knowledge} & 2018 & MCQ & Hybrid & KB & KB & ML \\
    
    \hline
    Bert with history answer embedding for conversational question answering  \cite{qu2019bert} & 2019 & Conv & Hybrid & RT & RT & DL \\

    \hline
    Framework for QA in Sanskrit through Automated Construction of Knowledge Graphs \cite{terdalkar-bhattacharya-2019-framework} & 2019 & MCQ & Fact & KB & KB & RB \\
    
    \hline
    Careful Selection of Knowledge to solve Open Book Question Answering \cite{banerjee2019careful} & 2020 & MCQ & Fact & RT & RT & DL \\
     
     \hline
     Open-Retrieval Conversational Question Answering \cite{qu2020open} & 2020 & Conv & RT & Hybrid & RT & DL \\
     
     \hline
     Bert representations for video question answering \cite{yang2020bert} & 2020 & Visual & Fact & Hybrid & Hybrid & DL \\
     
     \hline
     Counterfactual Samples Synthesizing for Robust Visual Question Answering \cite{Chen_2020_CVPR}
     & 2020 & Visual & Fact & Hybrid & Hybrid & DL \\
     \hline
     
    Decomposing visual question answering deep network via tensor decomposition and regression \cite{bai2021decomvqanet}
    & 2020 & Visual & Hybrid & Hybrid & Hybrid & DL \\
    \hline
    
    Graph-Based Reasoning over Heterogeneous External Knowledge for Commonsense QA \cite{Lv_Guo_Xu_Tang_Duan_Gong_Shou_Jiang_Cao_Hu_2020} & 2020 & MCQ & Fact & RT,KB & KB & DL \\
    
    \hline
    Research on Medical Question Answering System Based on Knowledge Graph\cite{9339902} & 2021 & MCQ & Fact & KB & KB & RB \\
    \hline

     \end{tabular}}
    \caption{Study of question type, answer type, evidence and answer sources, modeling approach for selected QA papers between the years 2017-2021. Full forms for the used short forms are given in \protect\footnotemark[2] }
    \label{table1} 
\end{table}

\footnotetext[2]{QT: Question Type, AT: Answer Type, ES: Evidence Source, AS: Answer Source, MA: Modeling Apporach, RC:Reading Comprehension, Conv:Conversational QA, Fact:Factoid QA, Def:Definition Based QA, RT:Raw Text, RB:Rule Based}

\subsection{Categorization based on type of question}
In this section the we are presenting major three directions of QA based on the type of questions, which includes multiple choice questions (MCQ) with the key challenge of eliminating the misleading options, conversational QA in which predominant dependency is on the conversation history and visual QA where evidence reference is derived from the farrago of text and visual features.

\subsubsection{MCQ Questions Answering}
Along with linguistic and relevant evidence retrieval challenges, answering the question with multiple options involves the challenge of equivocal reference\cite{nicula2018improving} where mapping candidate answers with evidence using standard IR approach can produce the misleading solution due to distraction of distorted options.

The technique of abduction is used by authors \cite{banerjee2019careful} by serving question hypothesis and domain knowledge to logical solver with the responsibility to abandon the possible misleading answers. Copynet Seq2Seq \cite{gu-etal-2016-incorporating} is adopted by \cite{banerjee2019careful} as a logic solver.

\subsubsection{Conversational QA}
\begin{figure}[h]
\centering
 \includegraphics[width=0.4\textwidth, keepaspectratio]{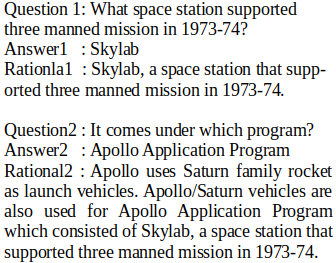}
 \caption{SQuAD example with the need of conversation history}
 \label{fig:CoQa_Ex_Squad} 
\end{figure}

Discourse plays important role in conversational question answering. Here the direct answering of question is not possible by looking at the context only as along with context the question depends on the conversational history \cite{qu2020open,braun2017evaluating,qu2019bert}. The conversation history plays vital role in answer generation. In the conversational question answering, the expected answer can be factoid, a definition or complex question either. The conversational question can be as shown in Figure \ref{fig:CoQa_Ex_Squad}.

By combining conversational response generation and reading comprehension problem, batter performance is observed by \cite{Reddy2019} in the conversational questions with free from text answers.

To handle conversational response generation problem, BiLSTM is used with encoder decoded in which at the encoder end passage, question-answer history is feeded as 
 ``p  q$_{i-n}$  a$_{i-n}$ ,,,,,q$_{i-1}$  a$_{i-1}$ q$_{i}$'' and decoder is responsible to generate the answer. As the answer words are likely to appear in the original passage, copy mechanism in the decoder is used. This model is referred to as the Pointer-Generator network, PGNet. Document Reader proposed by \cite{chen-etal-2017-reading} is used to handle the reading comprehension problem by predicting the rationale for given question.

\subsubsection{Visual Question Answering}
\begin{figure*}[h]
\centering
 \includegraphics[width=0.8\textwidth, keepaspectratio]{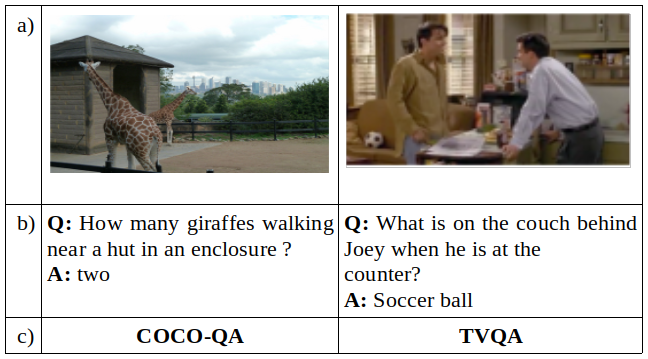}
 \caption{a) Example image\/frame b) question\-answer pair based on the visual content c) dataset from which example is derived}
 \label{fig:vqa_examples} 
\end{figure*}

Answering a question from visual content is referred to as Visual Question Answering (VQA) which embroil natural language processing and computer vision techniques \cite{yang2020bert,DBLP:journals/corr/RenKZ15}. Here, the visual content can be either image or video and the expected answer can be of factoid, non factoid, yes/no or any other types. The most recent approaches of VQA domain\cite{Chen_2020_CVPR} involve the use of a convolution neural network and a recurrent neural network to map the image and question features to common feature space. As per authors \cite{DBLP:journals/corr/WuTWSDH16}, the object identification, counting, or appearance, interactions, social relationships, or inferences about why or how something is occurring, are some example features which can be retrieved from visual content\cite{DBLP:journals/corr/Gupta17}. Figure \ref{fig:vqa_examples} indicates examples of visual question answering derived from the image dataset COCOQA\cite{10.5555/2969442.2969570} and TVQA\cite{lei-etal-2018-tvqa}, a dataset for video based question answering.

The model proposed by \cite{DBLP:journals/corr/AntolALMBZP15} uses LSTM to encode the question text and VGGNet\cite{DBLP:journals/corr/SimonyanZ14a} to encode the reference image. Features derived are combined after normalization and passed to a fully connected network. A softmax layer at the end obtains a distribution over answers.

\begin{itemize}
  \item \textbf{Video based Question Answering}
 \end{itemize}

Focus of \cite{lei-etal-2018-tvqa} is on visual question answering, in which to generate the feature set for network, visual features are used along with semantic features. To extract video features Faster R-CNN and ResNet101 is used. Bi-directional LSTM (BiLSTM) is used by author to encode both textual and visual sequences. A subtitle is flattened into a long sequence of words and GloVe is used to embed the words. Along with visual features, character name and subtitles are also specified to revamp the performance\cite{bai2021decomvqanet}.

Authors \cite{8419716} uses the datasets YouTube-QA, annotated video clip data \cite{DBLP:journals/corr/YuWHYX15} and TGIF-QA, annotated GIF data \cite{DBLP:journals/corr/LiSCTGJL16} for the video question answering system. The hierarchical parse tree based representation of question with the attention model\cite{yu2019deep} of video frames is core idea proposed by them. The model is trained on TreeLSTM which is diminished to a classification problem by adding softmax layer. The most promising answer from the candidate set is selected by the model. The attention input is designed from the input video frames using BiLSTM. In the parse tree, the leaf nodes represent the words of the question. The state of the leaf nodes comes from the word embedding\cite{ESPOSITO202088} and attention input to the important words. 

An approach to automatically harvest videos with description, candidate question answering pair generation and train the model based on modified existing approaches, is proposed by \cite{10.5555/3298023.3298196}. 
The focus of \cite{6ef97fabdf0a490884faa83519d7d596} is to design end to end memory network with the data as, statements followed by question with single word answer. This architecture is modified by \cite{10.5555/3298023.3298196} by replacing the statements (sequence of words) by video (sequence of frames). 
The bi-directional LSTM is used to encode the sequence of frame representation. Visual Question Answering model proposed by \cite{DBLP:journals/corr/AntolALMBZP15} is for the question answering for a single image. Here, in the modified approach \cite{10.5555/3298023.3298196} sequence of frames and the questions are encoded by two separate LSTM. To join both result element wise multiplication is performed.

In the soft attention model \cite{yao2015describing}, the dynamic soft attention\cite{xiong2016dynamic} is applied to frames. In the modified version \cite{10.5555/3298023.3298196}, the question is also encoded to while paying the attention to the frame.

The encoder decoder architecture of \cite{7410872} learns to encode the video and decode the sentence. video and question are encoded followed by decoding of answer in the modified version of \cite{10.5555/3298023.3298196}. 

\subsection{Categorization based on Answer Type}
One way to classify current research of question answering filed is, based on the types of expected answer \cite{yang2019end,10.1007/978-3-319-76941-7_72,shwartz2020unsupervised,1162-natural,chen-etal-2017-reading}which is generally further divided into factoid, definition, or complex question answering. 

\subsubsection{Factoid Questions Answering}
Factoid questions are the ones that ask about a simple fact and can be answered in a few words. A question \textit{"Which is the Capital of India?"} can be answered by single word \textit{"Delhi"}, is an example of factoid question. 

The task of answering knowledge based factoid question rely on inference, requires the aggregation of information from multiple sources. Automatic aggregation often fails because it combines semantically unrelated facts leading to bad inferences. The works by\cite{10.1007/978-3-319-76941-7_72,shwartz2020unsupervised} address this inference drift problem using modified page ranking and develops unsupervised and supervised mechanisms to control random walks on Open Information Extraction(OIE) knowledge graphs. 

\subsubsection{Definition based Question Answering}
In the definition based question answering \cite{1162-natural,chen-etal-2017-reading} the aim is to derive the short passage from the knowledge base, which satisfies the question constrain.  A question - answer pair \textit{"Q: What is Machine Learning?"}, \textit{"A: Machine learning is a field which provides the ability to learn the knowledge automatically, to computer systems."} is an example of definition based question answering.

In the research work authors \cite{1162-natural} uses Wikipedia pages to generate training model of short and long answers. Generation of dataset involves, annotator with a question, Wikipedia page pair. The annotator returns a (long answer, short answer) pair. Wikipedia articles are used to answer the question by authors \cite{chen-etal-2017-reading}. Bigram hashing with TF-IDF matching is used to train multi layer recurrent neural network. The trained model can detect the answer from Wikipedia paragraph. 

\begin{figure*}
\centering
 \includegraphics[width=0.8\textwidth,keepaspectratio]{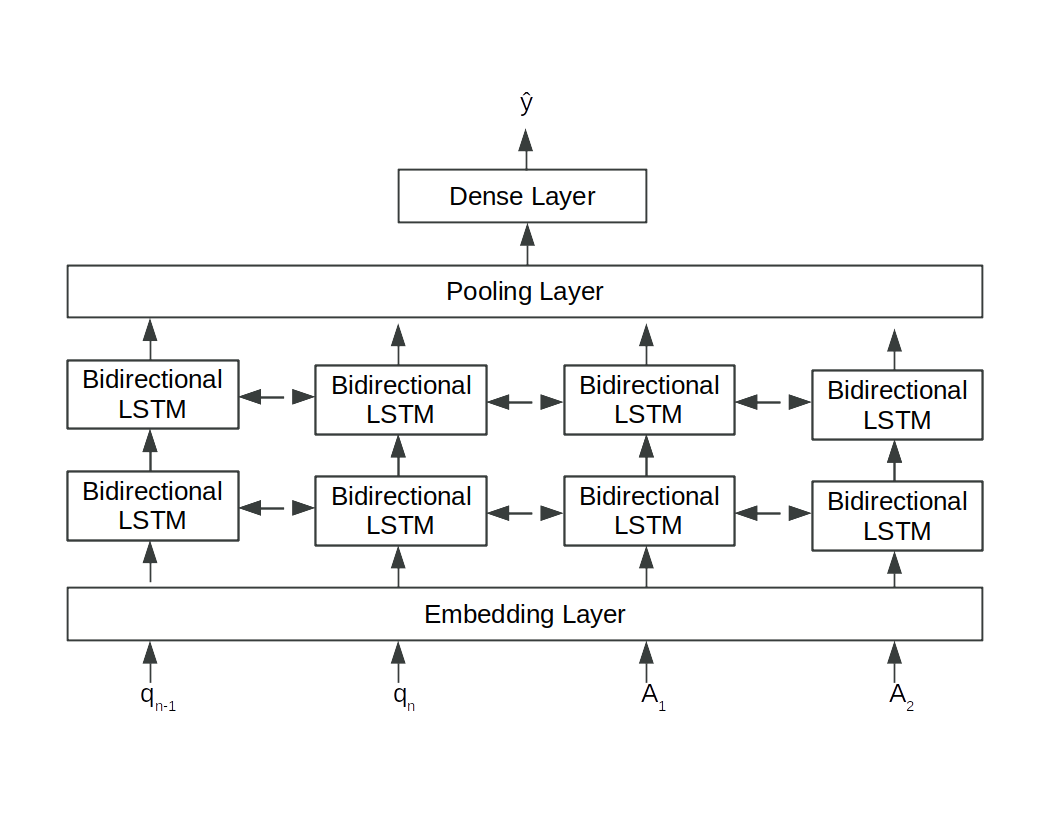}
 \caption{Bidirectional LSTM with n length question and answer text}
 \label{fig:BLSTM_QA} 
\end{figure*}

The authors \cite{10.1145/2970398.2970438} approach the task using deep learning methods without the need of feature extraction. The end to end training is achieved with a Bidirectional Long Short Term Memory (BLSTM) network and embedding layer is also involved in the training finetune the word representation as shown in the Figure \ref{fig:BLSTM_QA} drawn with reference to \cite{10.1145/2970398.2970438}.

\subsubsection{Hybrid Question Answering}
Answering the hybrid question like: \textit{Which is the second largest planet in the solar system?} involves many natural language challenges \cite{yang2019end}. Answer to such question can be factoid or non factoid, but the major challenge here is the selecting statements from the knowledge base/graph to satisfy the question statements. we need to use several semantic clues like : the answer is a planet, the answer is contained by solar system, the answer ranks second, the shorting order is descending order ,etc. 

Multiple predicates are required to constrain the answer set for such questions complex questions, if the knowledge graph is used \cite{huang2019knowledge,li2021incremental,bauer-etal-2018-commonsense}. To tackle this problem Luo et al. \cite{luo2018knowledge}, first generate multiple query graphs for a given question. Next semantic similarities between the question and each query graph using deep neural networks is calculated and final result from each candidate graph is combined for predicting the correct answer.

Approach of \cite{10.1145/3357384.3358026} divides the model in two parts: question interpretation, and answer inference. In the question interpretation phase the sets of entities and predicates that are relevant for answering the input question is identified with confidence score. In the second phase these confidence scores are propagated and aggregated over the structure of the knowledge Graph, to provide a confidence distribution over the set of possible answers.

\subsection{Categorization based on evidence or answer source}
Information can be stored as raw text or can also be represented as a knowledge graph. Based on the storage where we are looking for the answer we can divide this type further as raw text based  QA and Knowledge based QA(KBQA).

\subsubsection{Raw text based Question Answering}
\begin{figure}[h]
\centering
 \includegraphics[width=0.4\textwidth, keepaspectratio]{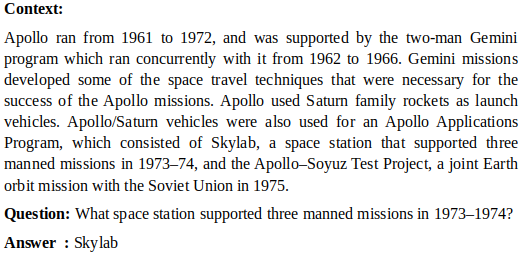}
 \caption{SQuAD:Context,Question-Answer example}
 \label{fig:squad_example} 
\end{figure}

The reasoning disparity between humans and machines is at a peak when it comes to answering a question from a passage. For a given question, humans can detect the answer paragraph or sentences from a given passage at ease while it is the challenge for the machine \cite{izacard2020leveraging,karpukhin2020dense}. An example of context, question and answer pair of SQuAD is mentioned in the Figure \ref{fig:squad_example} to represent this context.

Much research in this direction is going on and still not able to beat the human performance. Datasets like SQuAD \cite{rajpurkar-etal-2016-squad,rajpurkar-etal-2018-know}, WikiQA \cite{yang-etal-2015-wikiqa}, MS Marco \cite{DBLP:journals/corr/NguyenRSGTMD16} ,TriviaQA \cite{joshi-etal-2017-triviaqa} and many more are designed to tackle this challenge and to improve the machine comprehensive performance by using deep neural networks techniques involves but not limited to Bi-LSTM, attention based learning, distinct supervise learning, BERT, GPT models etc. 

Authors \cite{clark-gardner-2018-simple} approach this issue on \cite{rajpurkar-etal-2016-squad, joshi-etal-2017-triviaqa} datasets. The method followed by them includes pre-trained Glove word embedding combined with character embedding where the character embedding is generated using CNN followed by max-pooling. Bi-directional GRU is used to map the above embedding to context aware embedding. Lateral question and context are merged using attention mechanism. Further using bi-directional GRU start and end token of probable answers are identified.

The model of \cite{chen-etal-2017-reading} is used to find the answer for given question using Wikipedia. The model is of two part 1) Document retriever : to find relevant articles 2) Machine comprehension model : document reader for extracting answer from small set of documents. In the Document retriever part, for a given question 5 Wikipedia articles are retrieved using TF-IDF and bigram matching methods. In the Document Reader part, for all paragraphs, Bi-LSTM is trained using features like 1) Glove word embedding, 2) exact matching of paragraph word pi with one question word, 3) POS,Named Entity Recognition and Term frequency of each paragraph words and 4) attention score of similarity between each question word qj and current paragraph word pi. Result of all hidden layers are concatenated to obtain final features of pi. For a given question, on the top of word embedding of qi RNN layer is added. Result of all hidden units are concatenated in single vector \{q1,q2,,,ql\}. Once we have vectors for paragraphs and question then for each pair the goal is to predict span of token that is most likely the correct answer. Paragraph vector and question vectors are taken to predict two ends of span using two separate classifiers.

\subsubsection{Knowledge based Question Answering}
\begin{figure}[h]
\centering
 \includegraphics[width=0.45\textwidth, keepaspectratio]{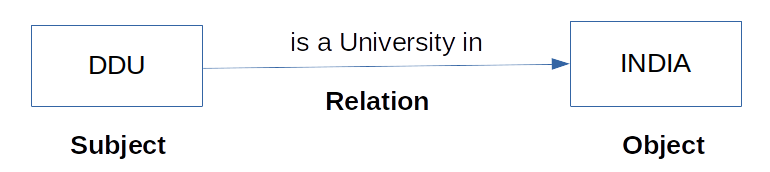}
 \caption{Knowledge Graph : Triplet Example}
 \label{fig:triplet_example} 
\end{figure}

The ontology based information represented is commonly refer as a knowledge graph. In the knowledge graph the information can be represented as [subject, relation, object] triplet, where \textit{relation} indicates the connection via which \textit{subject} and \textit{object} are connected. One example of such triplet is shown in the Figure \ref{fig:triplet_example}. Here, the relation \textit{'is a university in'} connects subject \textit{'DDU'} and object \textit{'INDIA'}.

Main challenge here is to map question words onto the graph and finding appropriate connections between nodes to locate the answer \cite{Lv_Guo_Xu_Tang_Duan_Gong_Shou_Jiang_Cao_Hu_2020,10.1007/978-3-030-19570-0_35}. Starting with simple mapping of question-answer to dividing the question into multiple subset and processing each with deep learning based model \cite{zhang2018variational,huang2019knowledge}, this direction has attracted many researchers.

\subsection{Categorization based on Modeling approach}
Starting with word matching model to recent transformer based models \cite{brown2020language,devlin2018bert} general classification of model is in rule based, machine learning based and deep learning based models. 

\subsubsection{Rule based Models}
In the majority of approaches of rule based models, based on WH question type different rules are designed\cite{riloff2000rule}. Language processing tasks like, semantic class tagging, and entity recognition plays a major role in such systems. A rule can be a logical combination of any of the above tasks. Each rule awards some points to a all sentences of the input \cite{7980526}. After all of the rules have been applied, the sentence that obtains the highest score is returned as the answer. In the current era mainly models for low resource languages are based on such rules \cite{terdalkar-bhattacharya-2019-framework} while others are diverted to machine learning or deep learning based system where significant performance improvement is observed.

\subsubsection{Machine Learning based models}
Majorly parsing result of question is given to machine learning models like support vector machine(SVM), decision tree(DT), naive bayes(NB) for the purpose of classification\cite{joachims2002learning,10.1007/978-3-030-19570-0_35,Lahbari2018TowardAN}. Other prediction problem using machine learning can be a prediction whether the given community question \cite{morris2010people} will entertained by some answer or not using predefined feature sets \cite{molino2016social,liu2011predicting}.
\subsubsection{Deep Learning based models}
With the availability of computing power at ease and an introduction of recurrent neural network(RNN) based models in the text processing field, the research progress of QA is diverted from pure machine learning based models to deep learning based models \cite{chen-etal-2017-reading,10.1145/3357384.3358026,Reddy2019,lei-etal-2018-tvqa}.

For a given question answering task, finetuning of pretrained transformer models like Google's BERT \cite{devlin2018bert} or OpenAI's GPT \cite{brown2020language} is the current state of art.

\section{Neoteric challenges of QA field}
Current advancement in the technology and eminent progress in deep learning leads question answering field beyond simple text based question and answer \cite{clark2020tydi} and thus have opened a door for new directions. The directions where notable progress is expressed in contemporary era includes semantic or discourse related challenges, automatic question generation, similarity detection between questions and challenges for low resource languages etc.

\subsection{Semantic related challenges}

Semantics, deals with the meaning of word or statement, is playing pivotal role in the understanding the knowledge structure, finding relationship among entities and objects and retrieval of evidence from ambiguous knowledge\cite{khurana2017natural}. Majority of QA systems are divided in two components named as  evidence generation and answer retrieval \cite{chen-etal-2017-reading,nicula2018improving}. To accomplish the evidence generation part in knowledge graph based QA systems generation of the sub-graph with relevant nodes is required while the evidence passage retrieval method is needed in the reading comprehension to achieve it. Semantic nets,first order predicate logic,conceptual graphs, conceptual dependency,  are some techniques used for finding the proper meaning of word.
 
\subsection{Discourse related challenges}

Discourse plays a vital role in the conversational question answering system \cite{chen2018dialogue} in which the current question depends not only on the evidence but also on the history of conversation.

Anaphora  Resolution- the method of deriving the correct meaning by detecting anaphora-antecedent  pairs\cite{voita2018context,poesio2018anaphora} and Discourse Structure Recognition - the arrangement of text structure \cite{huber2019predicting}, are two major directions which comes under discourse level challenges.

\subsection{Automatic question generation}

The task of automatically creating questions that can be answered by a certain span of text within a given passage, is explored by \cite{zhao-etal-2018-paragraph}. This involve the challenge of natural language generation (NLG) \cite{lewis2019bart}. Long text has posed challenges for sequence to sequence neural models in question generation – worse performances were reported if using the whole paragraph (with multiple sentences) as the input \cite{du2017learning}. The maxout pointer mechanism with gated self-attention encoder is proposed  to address the challenges of processing long text inputs for question generation. In the model design at the encoder side recurrent neural network (RNN)\cite{indurthi2017generating} is used and in the decoding stage, the decoder is another RNN that generates words sequentially conditioned on the encoded input representation and the previously decoded words.

Entity Relationship triplet from the knowledge graph is extracted by authors \cite{reddy-etal-2017-generating}. These keywords are given to recurrent neural network to generate question from that sequence. Answer to that question will be one of that keyword.

\subsection{Question Similarity Detection}

The direction of finding question similarities involves challenge of few words when it come to detect the similarity. For example, words used for framing the questions might different but still have semantic, like \textit{'Is it a watchable movie?'} and \textit{'Shall we plan for that movie?'} should identified as similar question. In opposition to that \textit{'Is it a watchable news channel?'} should identified as different question even it shares most of the words and language model of that words.

Especially after removing the stop words, the vector of each question is of very limited words. Finding the similarity between them involves advance level processing instead of applying only the methods like TF-IDF similarity, Resnik similarity, n-gram matching and other similarity measures and thus it attracts many researcher\cite{haponchyk-etal-2018-supervised,DBLP:journals/corr/abs-1808-08357}. Supervised clustering of questions into users intent categories, by using semantic classifiers is the focus direction of \cite{haponchyk-etal-2018-supervised}.

Approach used by \cite{DBLP:journals/corr/abs-1808-08357} is like, for a given question POS tagging followed by stop words removal is applied to identify keywords and that 
are used in tf-idf based ranking of all questions of dataset. Next top n results are compared with the user query for finding similarity scores.

\subsection{Challenges related to low resource languages}

Different natural languages behold different levels of progress in this field of question answering. For example, the aim of \cite{terdalkar-bhattacharya-2019-framework} is to automatic generation of the knowledge graph in the form of a triplet which indicates human relationships (like father, mother, sister, etc.) from Mahabharata.  Input question consists of Sanskrit text(in digital Unicode format)  and type of human relationship intended. The output is a set of triplets in the form [subject,
predicate, object] where the predicate is of the relationship type intended and subject and object are entities. If [a, R, b] is an output triplet, then it implies that object b is relation R of a subject a. 

As the focused natural language is Sanskrit, many pre-processing tasks are available to some limited extends only.  Pre-processing and knowledge base generation from raw data are major challenges that need to handle by a researcher working with less popular or less explored natural languages \cite{al2017aroma,karakanta2018neural,adams2017cross,gu2018meta,7980526}. Many Indian languages are testaments for this argument. Authors \cite{bapat2010paradigm,jena2011developing,goyal2008hindi,chen2020improved} developed morphological analyzer using paradigm based approach. For resolving ambiguities among the rules, dictionary based methods and statistical methods are used \cite{khalifa2017morphological}. Major challenge of this work is hand crafted rule building for a language that is highly inflectional with limited availability of the resources. Cross-lingual learning techniques are gaining popularity to tackle the resource constrain issue of low resource languages \cite{lample2019cross,hu2020xtreme,guan2020cross,wu2021reinforced,ocquaye2021cross}.

\section{Question Answering Popular Datasets}

\begin{figure*}[h]
\centering
 \includegraphics[width=0.9\textwidth,keepaspectratio]{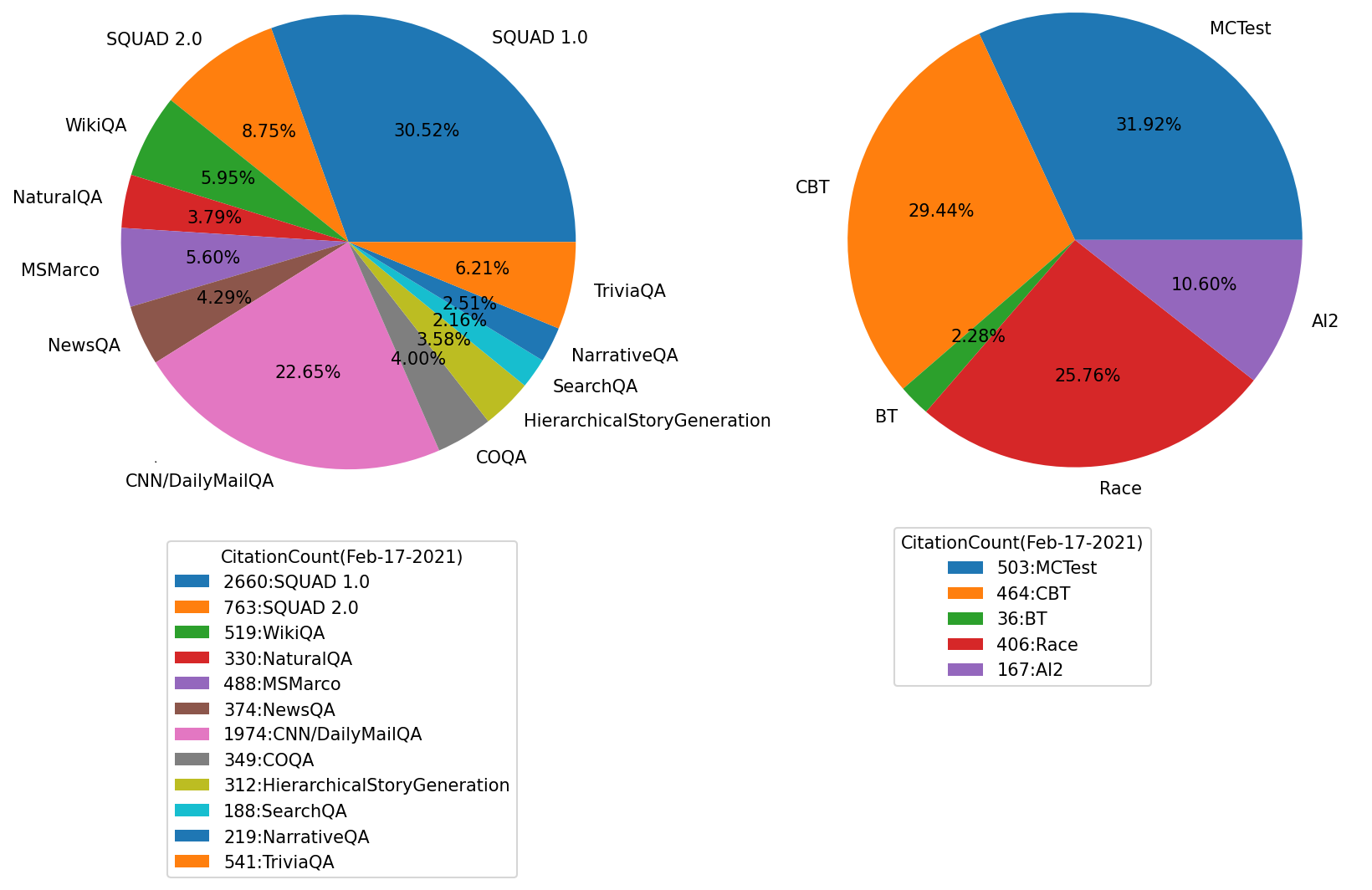}
 \caption{Dataset Citation Count}
 \label{fig:citeCount} 
\end{figure*}

In the field of question answering a gigantic sequence of questions and answers produced by a human is necessary to train a network. This can be accomplished using  largely available datasets which are in the form of row text documents or in the form of graph based structure referred as the knowledge graph. Mostly all the datasets are generated either by crowd-sourcing or by manual annotation. In this survey, some of the ferociously used datasets for machine reading comprehensive for single answer and multiple choice question answering(MCQ), are highlighted. Figure \ref{fig:citeCount} indicates Google Scholar\footnote[3]{https://scholar.google.com/} citation count of referred dataset papers till the date 17-Feb-2021.

\subsection{Machine Comprehensive Datasets}
\begin{itemize}
\item \textbf{Stanford Question Answering Dataset (SQuAD)}\\
Stanford Question Answering Dataset (SQuAD) is a reading comprehension dataset. It is available in two versions\cite{rajpurkar-etal-2016-squad,rajpurkar-etal-2018-know} SQuAD1.1 and SQuAD2.0. Here, questions and answering pairs are created through crowd-sourcing by considering Wikipedia as a base. SQuAD 1.1, contains 107,785 question-answer pairs on 536 articles. SQuAD2.0 combines question-answering pairs of SQuAD1.1 with 53,775 unanswerable questions from same paragraphs.

\item \textbf{WikiQA}\\
The WikiQA dataset \cite{yang-etal-2015-wikiqa} contains 3,047 questions by extracting queries from Bing query logs, using simple heuristics like starts with WH question and end with '?'. To identify the answer topic, the summary paragraph of the Wikipedia page related to a user query is retrieved. As the summary section of the Wikipedia page usually provides important information, sentences of it are assumed as candidates for the answer. A question with extracted paragraph is crowd-sourced to identify answer sentences. 

\item \textbf{Natural Questions}\\
In \cite{1162-natural}, questions consist of queries submitted to Google search engine. Along with a question, the Wikipedia page from the top 5 search results is used to find candidate answer. A long answer (typically a paragraph) and a short answer (one or more entities) if present in the Wikipedia page is supplied. The public release consists of 307,373 training examples with single annotations, 7,830 examples with 5-way annotations data for training and 7,842 for testing.

\item \textbf{MS Marco}\\
The dataset \cite{DBLP:journals/corr/NguyenRSGTMD16} contains 10,10,916 questions sampled from Bing’s search query logs with a human-generated answer and 1,82,669 human rewritten generated answers.  It also contains 88,41,823 passages, extracted from 35,63,535 web documents. The dataset contains Questions, types of question, answers, well formed answer and document from which answer passage is extracted.

\item \textbf{NewsQA}\\
NewsQA \cite{trischler-etal-2017-newsqa} contains 119,633 natural language questions posed by crowd-workers on 12,744 news articles from CNN. Text span of article, which contains answer is highlighted  by crowd-workers. For 1000 randomly selected examples of NewsQA and SQuAD, to identify the reasoning category, manual labeling is applied in one of the five categories as 1)Word Matching, 2) Paraphrasing, 3) Inference, 4)Synthesis, 5)Ambiguous/Insufficient.

\item \textbf{CNN/DAILY MAIL}\\
In paper \cite{10.5555/2969239.2969428} almost 93k articles from the CNN and 220k articles from the Daily Mail, are scraped with corresponding questions. Questions are constructed synthetically by deleting a single entity from summary of each article.  To identify the correct answer involves recognition of textual entailment between the article and the question. The named entities within an article are identified in a pre-processing step.

\item \textbf{CoQA}\\
Dataset \cite{Reddy2019} contains 127k question-answering pair, obtained from 8,000 conversations of text passages. Crowd-workers are grouped in two: questioners and answerers. For a given question, the questioner asks questions for which answerer submit a reply. Both can see the ongoing conversation during this process. The passages are from seven includes children’s stories from MCTest \cite{richardson-etal-2013-mctest}, literature from Project Gutenberg, 5 middle and high school English exams from RACE \cite{lai-etal-2017-race}, news articles from CNN \cite{10.5555/2969239.2969428}, articles from Wikipedia, Reddit articles from the Writing Prompts dataset \cite{fan-etal-2018-hierarchical}, and science articles from AI2 Science
Questions. For each in-domain dataset, 100 passages are kept for development and 100 for the test set, remaining passages are supplied as a training set.

\item \textbf{SearchQA}\\
SearchQA, \cite{DBLP:journals/corr/DunnSHGCC17} consists of more than  140k  question-answer pairs with each pair having 49.6 snippets on average. Each question-answer-context tuple of the  SearchQA comes with additional meta-data such as the snippet’s URL which can be valuable resources for future research as per the authors. Set of question-answer pairs are crawled from J! Archive, which has archived all the question-answer pairs from the quiz competition television show Jeopardy!. 
 
\item \textbf{NarrativeQA}\\
In NarrativeQA \cite{kovcisky2018narrativeqa} dataset total 1,567 stories are the collection from 1) books: fetched from Project Gutenberg and 2) movie scripts: scraped from the web. Full data is  partitioned into non-overlapping training, validation, and test portions, along with stories/summaries.  The dataset contains 46,765 question-answer pairs written by human annotators. Most of the questions are form WH with approx 9.8 tokens. 

\item \textbf{TriviaQA}\\
TriviaQA \cite{joshi-etal-2017-triviaqa} is the collection of total 650k question, answer and evidence triple. Most of the question often requires multi-sentence reasoning and retrieved from 14 trivia and quiz-league websites. For retrieval of the evidence two approaches are used:1) Wikipedia articles,2) Using top 50 web search result using Bing web search API.

\end{itemize}

\subsection{MCQ Datasets}

\begin{itemize}

\item \textbf{MCTEST}\\
To collect the dataset \cite{richardson-etal-2013-mctest}, crowd-sourced workers were instructed to write a short children story and four multiple-choice questions-answers for that story. Workers were supposed to generate option set such that all option words are from the story. At least two questions are such that it requires multiple sentences from the story to answer. The MCTest corpus contains two sets of stories, named MC160 of 160 stories and MC500 of 500 stories. Grammar errors done by a worker are manually corrected in MC160 while in MC500 writing task is assigned after taking grammar test.

\item \textbf{CHILDREN’S BOOK TEST}\\
The Children's book test (CBT) \cite{DBLP:journals/corr/HillBCW15} is built from freely available 108 books. Total 6,69,334 training 8,000 validation and 10,000 question are developed. To design question-answer pairs, initial 20 sentences from the context, and expected answer word (which must be one of four POS class Named Entities, Noun, Verb or Preposition), is removed from the 21 st sentence, which becomes the query. Models must identify the answer word among a selection of 10 candidate answers appearing in the context sentences and the -query.

\item \textbf{BOOK TEST}\\
Approach to collect the dataset \cite{DBLP:journals/corr/BajgarKK16} is similar to CBT \cite{DBLP:journals/corr/HillBCW15}.  Authors identifies whether each sentence contains either a named
entity(NE) or a common noun(CN) that already appeared in one of the preceding twenty sentences. If so, it is then replaced by a blank question. The preceding 20 sentences are used as the context document. Total 3555 copyright-free books to extract CN examples and 10507 books for named entity NE examples.
The dataset contains 14, 140, 825 training examples and 7, 917, 523, 807 tokens. The validation and test set consists of 10, 000 NE and 10, 000 CN questions each. The training data consist of tuples (q, d, a, A),  where q is a question, d is a document that contains the answer to question q, A is a set of possible answers and a $\in$ A is the ground truth answer. Both q and d are sequences of words from the vocabulary.

\item \textbf{RACE}\\
RACE \cite{lai-etal-2017-race}, consists of 27,933 passages and 97,687 questions of English exams for middle-school and high-school Chinese students. Each MCQ question is associated with four candidate options, one out of them is correct. Questions and candidate answers are free from text spans of the passage.

\item \textbf{AI2 Reasoning Challenge (ARC)}\\
The ARC dataset \cite{DBLP:journals/corr/abs-1803-05457} consists of 7787 multiple choice non-diagram science questions. The dataset is divided into two parts,  a Challenge Set of 2590 “hard” questions and an Easy Set of 5197 questions. The question vocabulary uses 6329 distinct words. The questions of Challenge Set are answered incorrectly by a retrieval-based algorithm and a word co-occurrence algorithm both. 

\end{itemize}

\section{Evaluation Metrics}

There are many evaluation methods available for the field of question answering. Depends on the problem addressed, different authors compare their performance with benchmark using different evaluation matrix \cite{Durmus_2020}. Starting with Precision, Recall and F1 score upto some benchmark evaluation like BLEU,ROUGE are covered in this section.

\begin{itemize}

\item \textbf{Precision, Recall,F1 Score}

For the MCQ based Question Answering, standard way to measure performance is to calculate precision, recall or F1 score as given by next equations.

\begin{align} 
Precision = \frac{True Positive}{Predicted Positive} \\
Recall =  \frac{True Positive}{Actual Positive} \\
F1 Score = 2 * \frac{Precision * Recall}{Precision +Recall}
\end{align}

 \item \textbf{BLEU\cite{10.3115/1073083.1073135}}
 
 It is the evaluation approach based on modified n-gram precision where candidates are scored based on count of n-grams that are also present in any reference. Score would be aggregated with Geometric Mean. In the calculation of BLUE-1, for each Candidate word of machine generated output find:

1) $W_{max}$: Maximum number of times a word occurs in any single reference translation .

2) $Count_{clip}$: Clipped word count is a min from  ($W_{max},W_{count}$). Where $W_{count}$, indicates count of candidate word in candidate output.

3) Add $Count_{clip}$ for all words and divide it by total words of candidates output.

\item \textbf{METEOR\cite{10.5555/1626355.1626389}}

METEOR (Metric for Evaluation of Translation with Explicit ORdering) is a metric originally designed for the evaluation of machine translation output with the aim of avoiding the limitations of BLEU. 

A term mapping can be thought of as a line between a unigram in candidate statement, and a unigram in reference statement with the constraints as every unigram in the candidate translation must map to zero or one unigram in the reference. The alignment is a set of mappings between unigrams and incrementally produced with a series of stages. 

Each stage consists of two phases: 1) Using external modules ((i) Exact Match module ii) Porter steam module iii) WN synonyms module. ), list all possible unigram mapping. 2) In the second phase, a large subset of this module is selected. Each stage maps unigrams that have not been mapped to any unigram in any of the preceding stages.  By default, the first stage uses the “exact” mapping module, the second the “porter stem” module and the third the “WN synonymy” module and it is feasible to change this sequence to produce a different result. For large segment common in both segment, n-gram matches are used to compute a penalty p for the alignment.

\item \textbf{ROUGE\cite{lin-2004-rouge}}

ROUGE is a set of metrics for evaluating automatic summarization of texts as well as machine translations and question answering.
ROUGE-1 considers unigram, ROUGE-2 bigrams and so on, ROUGE-N measure ngrams.  To calculate ROUGE score precision, recall and F Score are measured for the overlapping ngrams. Recall refers to how much of the reference summary the system summary is recovering or capturing. Precision indicates how much of the system summary was, in fact, relevant. Variants of ROUGE are ROUGE-N, ROUGE-S, and ROUGE-L. 

\item \textbf{Word Mover's Distance(WMD) \cite{10.5555/3045118.3045221}}
To find the distance between two documents the word mover's distance is hyper-parameter free, interpretable and high retrieval accurate method.  The WMD distance measures the dissimilarity between two text documents as the minimum amount of distance that the embedded words of one document need to “travel” to reach the embedded words of another document.

\item \textbf{Sentence Mover’s Similarity (SMS)\cite{clark-etal-2019-sentence}}

This techniques is modified WMD for evaluating multi-sentence texts by basing the score on sentence embeddings. Computation is same to WMD except each document is represented as a bag of sentence embeddings rather than a bag
of word embeddings.

\item \textbf{BERT Score \cite{DBLP:journals/corr/abs-1904-09675}} 

Computes a score by leveraging contextualized word representations, allowing it to go beyond exact match and capture paraphrases better.

An alignment is computed between candidate
and reference words by computing pairwise cosine
similarity. This alignment is then aggregated in to precision and recall scores before being aggregated into a (modified) F1 score that is weighted using inverse-document-frequency values.

 \item \textbf{Mean Reciprocal Rank (MRR)\cite{voorhees1999trec}}
 
 In the document retrieval field, MRR is used to identify the ability of system in finding the correct answer. The Reciprocal Rank (RR) is the inverse of rank of first correct answer for individual question. i.e. for a given question i, it's rank ($rank_i$) is k, from all n answers generated by the system if kth answer is correct. Average of RR score of each question is considered as MRR.
 \begin{equation}
 MRR= \frac{1}{|Q|} \sum_{i=1}^{|Q|} \frac{1}{rank_i}
\end{equation}
 
\end{itemize}
\section{Observations}
In the recent era, human performance has been surpassed by many state-of-the-art models for pure factoid-based QA and thus, specific branches of QA involving conversational QA, visual QA, multi-hop QA, open-domain QA are gaining the attention of researchers. With the availability of sizeable datasets like SQUAD in addition to the support from hardware advancements in terms of high computing capability, deep Learning based techniques like transformers, RNN, attention mechanism are giving promising results over rule-based and machine learning-based approaches and hence, are part of the major state of the art models. The presence of pre-trained models like BERT, GPT, XLnet, DistilBERT, Text-to-Text Transfer Transformer(T5), UniLM, and Reformer, has changed the model designing approach to the fine-tuning of pretrained models over the training from scratch.

Table \ref{table2} indicates the result obtained by top-performing state-of-the-art models on popular QA datasets. The evaluation approach used by a dataset depends on the problem at hand and the representation of information in it thus, we have mentioned evaluation parameter with result. Additionally F1 score is also mentioned, if present otherwise the field related to that is kept blank in the table.
\begin{table}[!htbp]
    \centering
    \resizebox{\columnwidth}{!}{%
    \begin{tabular}{l|p{0.48\textwidth}|l|l}
     \hline
     \textbf{Dataset} & \textbf{Model} & 
    \textbf{Result} & \textbf{F1 Score} \\
    \hline

     \multirow{2}{*} {SQuAD 2.0}	& Human performance	& 86.831(EM)	& 89.452 \\ 
     & SA-Net on Albert (ensemble)
&	90.724(EM)	& 93.011 \\
     \hline

	\multirow{2}{*}{SQuAD 1.1}	& Human performance	& 82.304(EM)	& 	91.221 \\ 
     & LUKE (single model)
 &	90.202(EM)	& 95.379 \\
     \hline
     
     \multirow{2}{*}{TVQA}	& Human performance	& 93.44(Val)	& ----- \\ 
     & ZGF (single model) &	70.29(Val)	& ----- \\
     \hline
     
     \multirow{2}{*}{TVQA}	& Human performance	& 89.61(val)	& 	 \\ 
     & STAGE (span) (single model) &	70.5(Val)	& -----  \\
     \hline
     
      \multirow{2}{*}{NaturalQA} 	& PoolingFormer(Long answer)
	& 0.62448(R@P=90) & 0.79823	 \\ 
	
     &  ReflectionNet(Short answer) &	0.35046(R@P=90)
	& 0.64114 \\
     \hline 
     
     MS Marco	& DML	& 0.47(MRR@100Dev)	& ----- \\
     \hline
     
     \multirow{2}{*}{COQA}	& Human performance	& 89.4 (InDomain)	& ----- \\ 
     & RoBERTa + AT + KD (ensemble) &	91.4(InDomain)
	& ----- \\
     \hline

     TriviaQA &  &	82.99(Full EM)
	& 87.18 \\
     \hline
     
     AI2 (ARC) &  UnifiedQA + ARC MC/DA + IR &	81.4(Accuracy)
	& ----- \\
     \hline
     
     MCTest & Parallel-Hierarchical on Sparse
 &	71\% (MC500 Test)

	& ----- \\
     \hline
     \end{tabular}}
    \caption{Performance of state-of-the-art models on QA datasets}
    \label{table2} 
\end{table}

The majority of the above-mentioned progress is witnessed by resource-rich languages like English or French. For low resource languages, cross-lingual or multilingual approaches are better alternatives over language-specific models due to the limited availability of datasets to design the deep learning-based model.

\section{Conclusion}

In this paper, we have presented a survey of recent state-of-the-art QA systems. Based on the type of questions different research directions are analyzed followed by open challenges like reading comprehension, conversational question answering, visual question answering, question generation, challenges for low resource language. Some approaches used by researchers are presented with detailing. Finally, a survey of question answering datasets and evaluation measures is presented. We have also provided the observation that the deep learning techniques are gaining popularity for resource-rich languages; it is far off for low-resource languages where still rule-based or machine learning approaches are prevailing.


\bibliographystyle{apalike} 
 \bibliography{cas-refs}





\end{document}